\documentclass[conference]{IEEEtran}

\usepackage{cite}
\usepackage{amsmath,amssymb,amsfonts}
\usepackage{graphicx}
\usepackage{siunitx}
\usepackage{booktabs}
\usepackage{multirow}
\usepackage{hyperref}
\usepackage{graphicx}
\usepackage[caption=false,font=footnotesize]{subfig}
\usepackage{amsmath,amssymb}
\usepackage{tikz}
\usepackage{verbatim}
\usetikzlibrary{shapes.geometric, arrows.meta, positioning, fit, backgrounds}
\hypersetup{hidelinks}

\begin{document}

\title{Fourier-Enhanced Recurrent Neural Networks\\
for Electrical Load Time Series Downscaling}

\author{
\IEEEauthorblockN{Qi Chen}
\IEEEauthorblockA{Department of Statistics\\
University of Chicago\\
Chicago, IL, USA\\
Email: qic0@uchicago.edu}
\and
\IEEEauthorblockN{Mihai Anitescu}
\IEEEauthorblockA{
Mathematics and Computer Science Division\\
Argonne National Laboratory and \\
Department of Statistics \\
University of Chicago\\
Email: anitescu@anl.gov
}
}

\maketitle

\begin{abstract}
We present a Fourier-enhanced recurrent neural network (RNN) for downscaling electrical loads. The model combines (i) a recurrent backbone driven by low-resolution inputs, (ii) explicit Fourier seasonal embeddings fused in latent space, and (iii) a self-attention layer that captures dependencies among high-resolution components within each period. Across four PJM territories, the approach yields RMSE lower and flatter horizon-wise than classical Prophet baselines (with and without seasonality/LAA) and than RNN ablations without attention or Fourier features.
\end{abstract}


\section{Introduction}

Energy policy and infrastructure investment decisions require an integrated system-wide perspective that captures the interdependencies of supply, conversion, and end-use sectors, as well as feedback from macroeconomic, technology-cost, and policy drivers. Many such energy modeling systems exist \cite{fattahi2020systemic}, of which the National Energy Modeling System (NEMS), developed by the U.S. Energy Information Administration (EIA) \cite{national1992national}, is widely used by policymakers and stakeholders for this very reason. However, as noted in the study of energy plant pollution studies provided by NEMS \cite{hobbs2010regions}, using temporally and spatially averaged data may significantly miss essential features and pricing signals. As was noted there, such approaches typically produce yearly or, at best, quarterly estimates of relevant variables, such as load. This would not be sufficient to understand the effect of investment and policy on reliability and resilience, which are much shorter-term phenomena. 

One of the crucial elements of understanding the interaction between multi-decade policies and trends, as currently provided by NEMS, and reliability and resilience concerns requires \textit{temporal downscaling :} mapping the demand projections from tools such as NEMS into much finer time-scale demand patterns, consistent with previous load dynamics. In turn this allows carrying out essential analyses such as unit commitment \cite{saravanan2013solution} and economic dispatch \cite{xia2010optimal}, that can lead to important reliability and resilience metrics, such as loss of load expectation (LOLE). 

To this end, we propose leveraging advanced machine learning techniques to produce downscaled electricity load data, which involves predicting the future distribution of load at hourly scales, conditional on daily or yearly data. We note that one could produce fine temporal load forecasts with other tools, such as dsgrid \cite{hale2018demand}, and then aggregate them.  Such an approach, however, requires many assumptions that lack the necessary data, and they would not necessarily be consistent with policy or strategic simulator output, such as NEMS. Having a data-driven approach to downscaling, while limited in many other ways, has the advantage of not requiring additional assumptions to produce it.  This study has two objectives: (1) to produce a tool for downscaling time series, focused on electricity demand, and (2) to identify features of machine learning models that lead to improved such models to assist other efforts where load modeling is sought. We also note that temporal downscaling is a feature that appears in pollution modeling \cite{hobbs2010regions} and is frequently used in environmental science \cite{michel2021climate}, so our approach could, in principle, be applied there as well. We also note that, while demand forecasting is a well-established area \cite{mystakidis2024energy}, it differs from downscaling, which focuses on the accuracy of predictions conditional on coarser temporal scales, the focus of this study.

\section{Related Work}
While the literature on time series forecasting is vast, there is less work on temporal downscaling, and particularly in the context of energy demand. However, many features of forecasting will be helpful here, such as multi-resolution and covariates, so that we will discuss statistical models in that context. 
Classical statistical methods such as ARIMA, exponential smoothing, and linear regression capture short-term autocorrelation and explicit seasonality but are limited in representing nonlinearity and multi-resolution dynamics. 
The Prophet model introduced an additive decomposition framework combining trend, seasonality, and holiday effects, offering interpretability and robustness to missing data ~\cite{taylor2018prophet}. 
However the model tends to underfit dynamic or irregular periodic patterns commonly observed in power load data.

Recurrent neural networks (RNNs) and their gated variants, such as GRU and LSTM, address nonlinear temporal dependencies by learning hidden-state dynamics from data ~\cite{cho2014gru}. Yet, standard architectures process information strictly in the forward temporal direction and do not provide the conditional function of downscaling. 
Recent attention-based extensions partially mitigate this limitation but often rely on global self-attention, which can be computationally expensive and less interpretable for structured time series.
Approaches such as temporal convolutional networks and recurrent–attention hybrids
capture dependencies within and across time scales, but either lack explicit treatment of seasonality
or treat it as an implicit pattern learned by the network, which can limit interpretability
and robustness when seasonal factors vary over long horizons.

In contrast, the proposed model integrates multi-scale information directly within the recurrent structure. The RNN takes the low-resolution input as the primary driver while incorporating a Fourier path that provides periodic components for seasonal variation. By combining these two sources, the subsequent self-convolution layer effectively measures intra-future dependencies, enabling the model to capture fine-scale temporal interactions while preserving global coherence.

\section{Methodology}

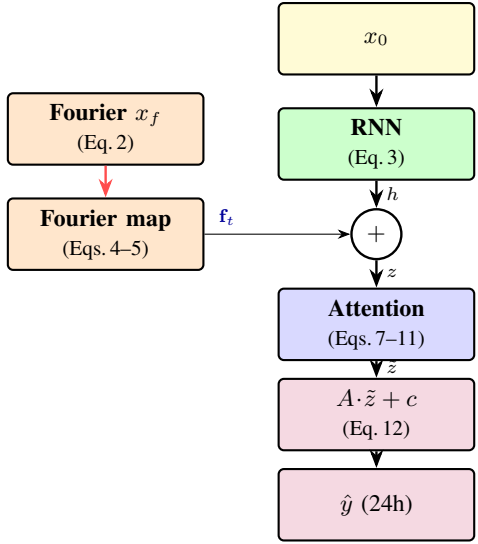
\begin{figure}[t]
\centering
\begin{tikzpicture}[
  font=\small,
  x=1cm, y=1cm,
  box/.style={draw, rounded corners=2pt, minimum width=2.6cm, minimum height=0.95cm,
              align=center, line width=0.8pt},
  inputbox/.style={box, fill=yellow!20},
  rnnbox/.style={box, fill=green!20},
  fourierbox/.style={box, fill=orange!20},
  attnbox/.style={box, fill=blue!15},
  outputbox/.style={box, fill=purple!15},
  sumcircle/.style={draw, circle, minimum size=0.6cm, line width=0.9pt, fill=white},
  arr/.style={-Stealth, line width=0.9pt},
  rarr/.style={-Stealth, line width=0.9pt, red!70},
  >=Stealth
]

\node[inputbox]  (x0)   at (0,6.6) {$x_0$};
\node[rnnbox]    (rnn)  at (0,5.2) {\textbf{RNN}\\[1pt]\footnotesize(Eq.\,3)};
\node[sumcircle] (plus) at (0,4.0) {$+$};
\node[attnbox]   (attn) at (0,2.8) {\textbf{Attention}\\[1pt]\footnotesize(Eqs.\,7--11)};
\node[outputbox] (head) at (0,1.6) {$A \!\cdot\! \tilde z + c$\\[1pt]\footnotesize(Eq.\,12)};
\node[outputbox] (yhat) at (0,0.4) {$\hat{y}$ (24h)};

\draw[arr] (x0)  -- (rnn);
\draw[arr] (rnn) -- node[right, pos=0.55, yshift=1pt] {\scriptsize $h$} (plus);
\draw[arr] (plus) -- node[right, pos=0.55, yshift=1pt] {\scriptsize $z$} (attn);
\draw[arr] (attn) -- node[right, pos=0.5,  yshift=1pt] {\scriptsize $\tilde z$} (head);
\draw[arr] (head) -- (yhat);

\node[fourierbox] (xf) at (-3.6,5.4) {\textbf{Fourier} $x_f$\\[1pt]\footnotesize(Eq.\,2)};
\node[fourierbox] (uf) at (-3.6,4.0) {\textbf{Fourier map} \\[1pt]\footnotesize(Eqs.\,4--5)};

\draw[rarr] (xf) -- (uf);
\draw[->] (uf.east) -- ++(0.6,0)
    node[midway, above, text=blue!60!black] {\scriptsize $\mathbf{f}_t$}
    |- (plus.west);

\end{tikzpicture}

\caption{Fourier-enhanced RNN architecture. Low-resolution path ($x_0$) drives the RNN; the seasonal path ($x_f{\to}f_t$) is fused at $z$, refined by a per-timestep self-attention block, and mapped to hourly output. \newline\scriptsize Eq.\,(3) RNN update; Eqs.\,(4--5) Fourier projection; Eqs.\,(7--11) self-attention; Eq.\,(12) output head.}
\label{fig:arch}
\end{figure}

\subsection{Inputs and Feature Construction}

We consider a multi-resolution time series composed of coarse observations
$\{x_{0,t}\}_{t=1}^{T}$ and their associated high-resolution vectors
$\mathbf{y}_t \in \mathbb{R}^{K}$, where $K$ denotes the number of
sub-periods within each period (e.g., hours within a day, or days within a week).
Each temporal instance $t$ therefore represents a multi-resolution pair
$(x_{0,t},\,\mathbf{y}_t)$ capturing both the aggregate signal and its internal
structure.

For each period $t$, the model receives a continuous input vector
\begin{equation}
\mathbf{x}_t = [\,x_{0,t};\ \mathbf{x}_{f,t}\,],
\end{equation}
where:
\begin{itemize}
    \item $x_{0,t} \in \mathbb{R}$ is a low-resolution scalar summarizing
    the aggregate magnitude (e.g., total demand or average value over the period);
    \item $\mathbf{x}_{f,t} \in \mathbb{R}^{2F}$ encodes \emph{Fourier seasonal features}
    that represent cyclic patterns across multiple temporal scales.
\end{itemize}

Let $t$ denote the period index measured along a continuous time axis at a low-resolution scale. Each high-resolution random variable $y_{t,k}$ within period $t$ is associated with a fractional index $t + k / K$, where $K$ is the number of sub-periods (e.g., hours within a day). Let $P>0$ represent the base cycle length (e.g., one week or one year)
expressed in the same units as $t$.
For a chosen number of harmonics $F$, we define the angular frequencies
$\omega_k = 2\pi k / P$ for $k = 1, \ldots, F$ \cite{bracewell1999fourier}. A within-period offset $s/K$ is introduced for each sub-period index
$s \in \{0, 1, \ldots, K{-}1\}$:
\begin{equation}
\label{eq:fourier-matrix}
\begin{aligned}
\mathbf{x}_{f,t}[s]
&=\big[
\sin\!\big(\omega_1 (t + \tfrac{s}{K})\big),\
\cos\!\big(\omega_1 (t + \tfrac{s}{K})\big),\ \ldots,\\[-2pt]
&\quad
\sin\!\big(\omega_F (t + \tfrac{s}{K})\big),\
\cos\!\big(\omega_F (t + \tfrac{s}{K})\big)
\big],
\end{aligned}
\end{equation}
Stacking \(\mathbf{x}_{f,t}[s]\) over all sub-periods produces
\(\mathbf{X}_{f,t}\in\mathbb{R}^{K\times 2F}\) for period \(t\).

\medskip
\noindent
This construction provides a smooth and differentiable encoding of periodic
structure, supporting both shared and sub-period-specific seasonal patterns
within the same modeling framework.

\subsection{RNN Backbone for Base Dynamics (acts on $x_0$ only) }
Let $\mathbf{h}_t \in \mathbb{R}^{L}$ denote the latent state ($L$ is the latent dimension).
We use a single-layer $\mathrm{RNN}$ (or $\mathrm{GRU}$ variant) that \emph{only} ingests $x_{0,t}$:
\begin{align}
\mathbf{h}_t &= \mathrm{RNN}\!\left(\mathbf{h}_{t-1},\, x_{0,t}\right),
\qquad \mathbf{h}_0 \text{ learnable.}
\end{align}
This isolates base-level dynamics in a compact recurrent path and preserves a distinct parameterization from the seasonality branch.

\subsection{Fourier Projection for Seasonality (acts on $\mathbf{x}_f$)}
Seasonal effects are introduced into the latent space through a separate, learnable projection. Each sub-period’s Fourier representation is compressed to the latent width 
using a shared per-row mapping $\mathbf{V} \in \mathbb{R}^{2F \times L}$:
\begin{equation}
    \mathbf{G}_t = \mathbf{X}_{f,t} \mathbf{V}, \quad \mathbf{G}_t \in \mathbb{R}^{K \times L}.
\end{equation}
The $K$ sub-period embeddings are then aggregated through a learnable 
softmax-normalized weight vector $\mathbf{a} \in \mathbb{R}^{K}$:
\begin{equation}
    \mathbf{f}_t = \mathbf{G}_t^\top \mathbf{a}, \quad \mathbf{f}_t \in \mathbb{R}^{L}.
\end{equation}
The resulting projection captures sub-period specific harmonic effects 
and is added to the recurrent latent path during fusion. The base and seasonal paths combine additively:
\begin{equation}
\mathbf{z}_t \;=\; \mathbf{h}_t + \mathbf{f}_t \;\in\; \mathbb{R}^{L}.
\label{eq:latent-fusion}
\end{equation}

\subsection{Self Attention over Latent Dimensions}
To model within-step correlations and incorporate future-known conditioning,
we apply self-attention over the \(L\) latent ``feature tokens’’ of each
\(\mathbf{z}_t\):
\begin{align}
\mathbf{T}_t &= \big[\mathbf{z}_t \odot E\big] \in \mathbb{R}^{L\times D}, \\
\mathbf{U}_t &= \mathrm{MHA}(\mathbf{T}_t,\mathbf{T}_t,\mathbf{T}_t), \\
\mathbf{H}_t &=\mathrm{LN}_1(\mathbf{T}_t + \mathbf{U}_t)\\
\Delta_t &= \mathrm{LN}_2(\mathbf{H}_t + \mathrm{FFN}(\mathbf{H}_t))\, \mathbf{w}_{\text{out}};\in \mathbb{R}^{L},\\
\tilde{\mathbf{z}}_t &= \mathbf{z}_t + \sigma\!\big(G\mathbf{z}_t\big)\odot \Delta_t ,
\end{align}

where $E\in\mathbb{R}^{L\times D}$ is a learnable feature embedding, $MHA$ is multi-head attention \cite{vaswani2017attention}; $LN$ is layer normalization \cite{ba2016layernorm};
FFN is a two-layer feed-forward network with ReLU and dropout mapping $\mathbb{R}^{D}\!\to\!\mathbb{R}^{D}$, $\mathbf{w}_{\text{out}}\in\mathbb{R}^{D}$ projects back to $L$,
$\sigma$ is a sigmoid gate, and $G\in\mathbb{R}^{L\times L}$ a learned gating matrix.

Under the downscaling setting, the self-attention layer enables the model to jointly encode amplitude, phase, and contextual effects across sub-periods within each forecast step. This design enhances the representation of fine-grained temporal structure and allows the model to effectively integrate known conditioning variables, improving intra-period coherence in the reconstructed high-resolution profile.

\subsection{Output Head}
The high-resolution vector predictions are linear in the attended latent state ( Fig.~\ref{fig:arch}, box “$A\cdot\tilde z + c$”; cf. (9) ):
\begin{equation}
\hat{\mathbf{y}}_t \;=\; A\,\tilde{\mathbf{z}}_t + \mathbf{c}, 
\qquad A\in\mathbb{R}^{K\times L},\; \mathbf{c}\in\mathbb{R}^{K}.
\end{equation}

Given the residual model $\mathbf{y}_t=\hat{\mathbf{y}}_t+\boldsymbol{\varepsilon}_t$ with
$\boldsymbol{\varepsilon}_t\sim\mathcal{N}(\mathbf{0},\boldsymbol{\Sigma})$,
the covariance matrix $\boldsymbol{\Sigma}$ is estimated as
\[
\boldsymbol{\hat\Sigma}
= \frac{1}{T-1} \sum_{t=1}^{T}
(\mathbf{r}_t - \bar{\mathbf{r}})
(\mathbf{r}_t - \bar{\mathbf{r}})^{\!\top},
\quad
\bar{\mathbf{r}} = \frac{1}{T}\sum_{t=1}^{T}\mathbf{r}_t.
\]
where $\mathbf{r}_t = \mathbf{y}_t - \hat{\mathbf{y}}_t$ represent the training residuals. Subsequently, this uncertainty model is used to produce confidence intervals that will be used for validation and hypothesis testing in \S \ref{ss:calibration}.

\section{Training Objective}
We train with mean-squared error (MSE) over sequences of length $T$:
\begin{equation}
\mathcal{L}_{\text{data}}
=
\frac{1}{T}\sum_{t=1}^{T}
\left\lVert \hat{\mathbf{y}}_t - \mathbf{y}_t \right\rVert_2^2 .
\end{equation}
To regularize seasonality, we penalize the Fourier projection $U_f$ with harmonic weights 
$w_k{=}k$ applied to the pair $(\sin, \cos)$ of order $k$:
\begin{equation}
\mathcal{L}_{\text{harm}}
=
\lambda_f \!\sum_{i,j} \big(U_f[i,j]\big)^{2} \, w_j^{2}.
\end{equation}
The overall objective is
\begin{equation}
\mathcal{L} \;=\; \mathcal{L}_{\text{data}} + \mathcal{L}_{\text{harm}}.
\end{equation}
We optimize $\mathcal{L}$ with Adam, apply gradient clipping, and forecast with known future features via a rolling window.

\section{Extended Framework for Yearly-to-Hourly Downscaling}

Building on the proposed RNN architecture, 
the framework is extended to the more challenging problem of mapping a yearly aggregate 
to a full hourly profile. Let $Y^{(Y)}$ denote the yearly aggregate (low resolution) 
and $Y^{(H)} \in \mathbb{R}^{365\times24}$ the target hourly field (high resolution). 
The proposed RNN structure can be applied in two ways to achieve the yearly-to-hourly 
downscaling task.

\subsection{Hierarchical RNN Downscaler}
The mapping could be decomposed into a hierarchical structure 
(e.g., year$\rightarrow$month$\rightarrow$day$\rightarrow$hour), 
where each stage is implemented with the proposed RNN-based model:
\begin{equation}
f_{\text{year}\rightarrow\text{hour}} 
= f_{\text{day}\rightarrow\text{hour}} 
\circ f_{\text{month}\rightarrow\text{day}} 
\circ f_{\text{year}\rightarrow\text{month}}.
\tag{15}
\end{equation}

The three-stage configuration above often exhibits a distinct monthly pattern, since the proposed RNN-based model assumes that the outputs are conditionally independent given the input. In contrast, a two-stage decomposition (year$\rightarrow$day, day$\rightarrow$hour) is adopted when the data exhibit less distinct monthly variation.

Except for the first stage, 
subsequent stages may be trained on a mixture of real and upstream-predicted inputs. 
After fitting $f_{\text{year}\rightarrow\text{month}}$, 
let $\widehat{Y}^{(M)} = f_{\text{year}\rightarrow\text{month}}(Y^{(Y)})$. 
A blended training input is then constructed as
\begin{equation}
\widetilde{Y}^{(M)} = 
\alpha Y^{(M)} + (1-\alpha)\widehat{Y}^{(M)}, 
\quad \alpha \in [0,1],
\tag{17}
\end{equation}
and used to train $f_{\text{month}\rightarrow\text{day}}$. 
The same strategy applies to $f_{\text{day}\rightarrow\text{hour}}$. 
This progressive scheme enables end-to-end deployment even when 
intermediate-resolution ground truth is sparse, 
while preserving cross-scale consistency inherited from the hierarchical formulation in equation (15).

\subsection{RNN-Enhanced Base Downscaler}

The proposed RNN-based model exhibits strong capability in modeling fine-grained temporal patterns and capturing intra-period variability. Another approach leverages this strength by employing the RNN module 
as a high-resolution refiner on top of a coarse-scale base downscaler. A base model 
(e.g., Prophet with linear accuracy adjustment (LAA) \cite{koutsoyiannis1996accurate}) first maps 
yearly aggregates to daily values:
\begin{equation}
\widehat{Y}^{(D)} = f_{\text{base}}\!\left(Y^{(Y)}\right),
\quad 
\widehat{Y}^{(D)} \in \mathbb{R}^{365}.
\tag{15}
\end{equation}
The RNN refiner then reconstructs hourly profiles conditionally on the 
daily signal:
\begin{equation}
\widehat{Y}^{(H)} = f_{\text{RNN}}\!\left(\widehat{Y}^{(D)}\right),
\quad 
\widehat{Y}^{(H)} \in \mathbb{R}^{365\times24}.
\tag{16}
\end{equation}
The resulting mapping is therefore the composition
\begin{equation}
f_{\text{year}\rightarrow\text{hour}} 
= f_{\text{RNN}} \circ f_{\text{base}},
\tag{17}
\end{equation}
where $f_{\text{base}}$ is any yearly to daily downscaler, and the proposed RNN-based 
module serves as a fine-resolution reconstructor. This formulation can 
also mitigate data scarcity in practice by leveraging reliable 
coarse-scale predictors while retaining high-resolution temporal detail.

\section{Experiments}

\subsection{Datasets}

We evaluate the proposed framework on four service territories from the
public PJM Hourly Energy Consumption dataset.
The selected regions are \emph{AEP}, \emph{COMED}, \emph{DAYTON}, and
\emph{DEOK}. Each dataset records hourly system load over multiple years. The selected training sets include a mix of data from both stable and upward-trending regimes,
thereby testing the model's ability to generalize across heterogeneous
load behaviors within a unified training context.

We remove duplicated timestamps, and forward-fill missing hours. Each dataset is split chronologically into training and test segments. All continuous variables are normalized using z-scores computed on the training partition.

\subsection{Models Compared}

Two comparisons are conducted to evaluate the proposed framework.
First, within the RNN structure, a Simple RNN shared the same RNN Backbone as the proposed model serves as a baseline that. A second variant, RNN with self-convolution, extends this baseline by introducing a self-convolution layer. Second, across modeling families, comparisons are made against the Facebook Prophet model under two configurations: with or without weekly, monthly, and yearly seasonality components, and with or without Linear Accuracy Adjustment (LAA). Given past hourly loads and exogenous features, we forecast the next 24 hours. We report horizon-wise $\text{RMSE}(h)$ for $h{=}0,\dots,23$.

\begin{figure}[t]
\centering
\subfloat[AEP]{\includegraphics[width=0.48\linewidth]{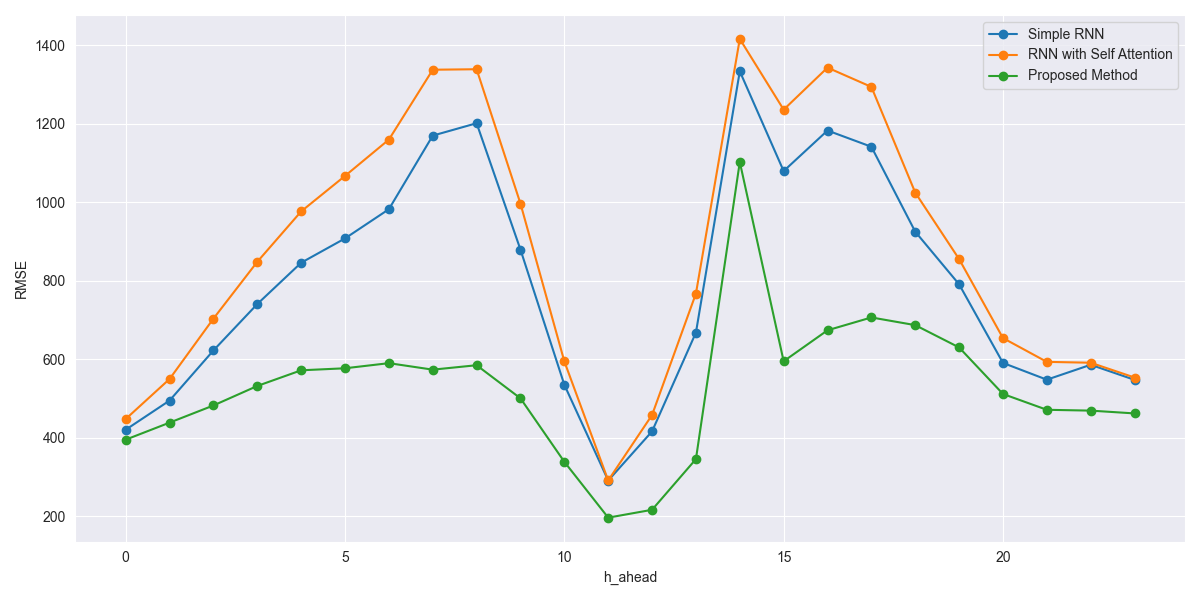}}
\subfloat[COMED]{\includegraphics[width=0.48\linewidth]{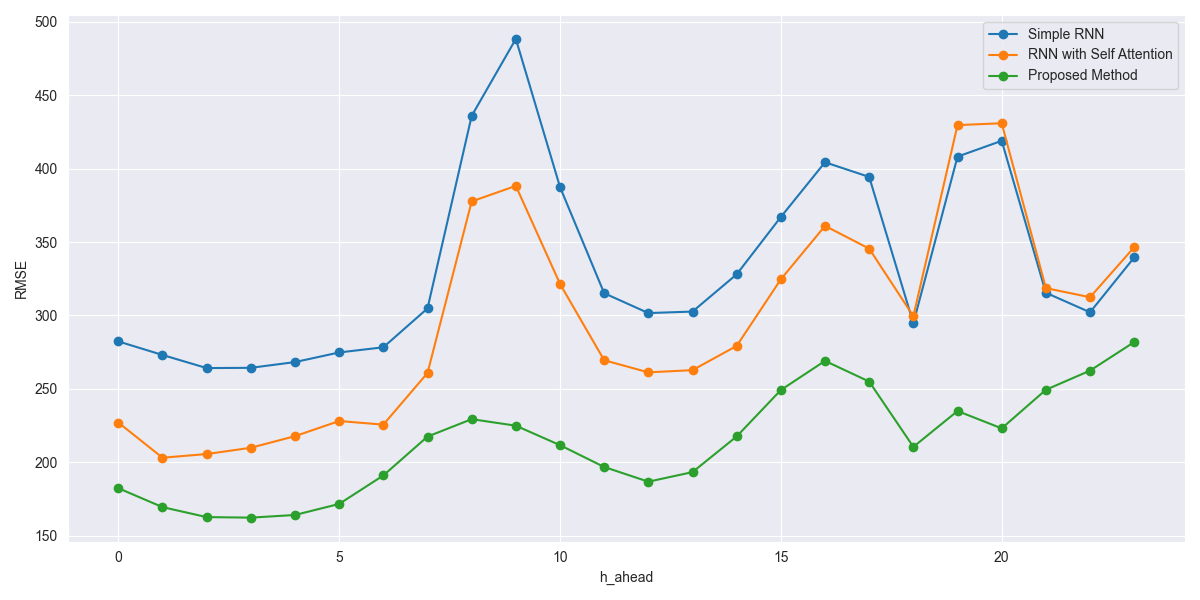}}\\[-2pt]
\subfloat[DAYTON]{\includegraphics[width=0.48\linewidth]{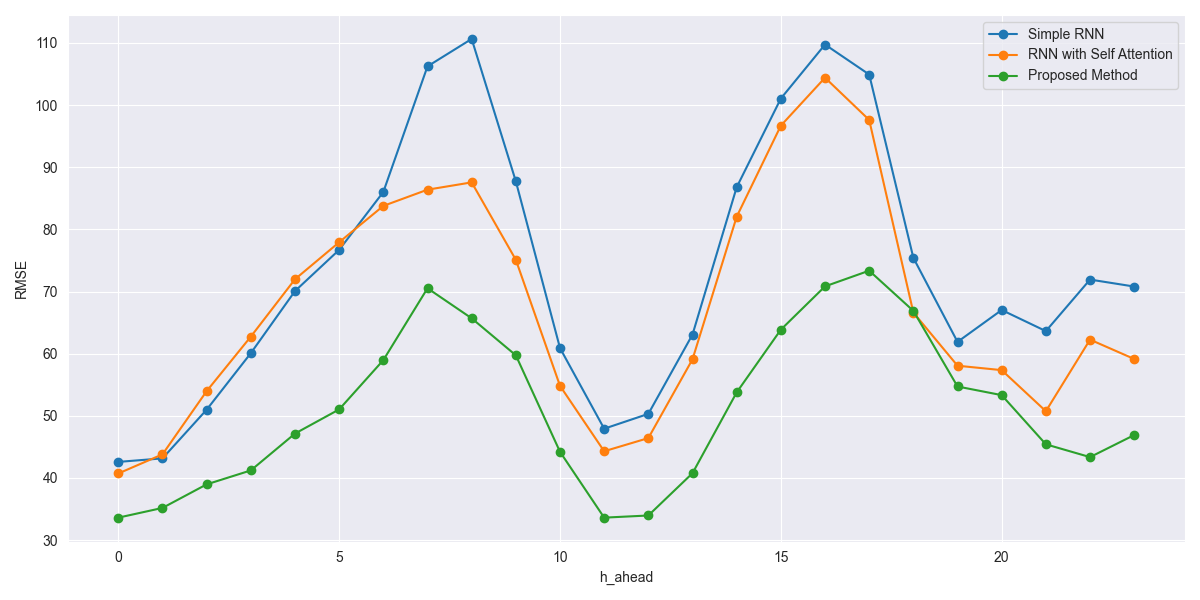}}
\subfloat[DEOK]{\includegraphics[width=0.48\linewidth]{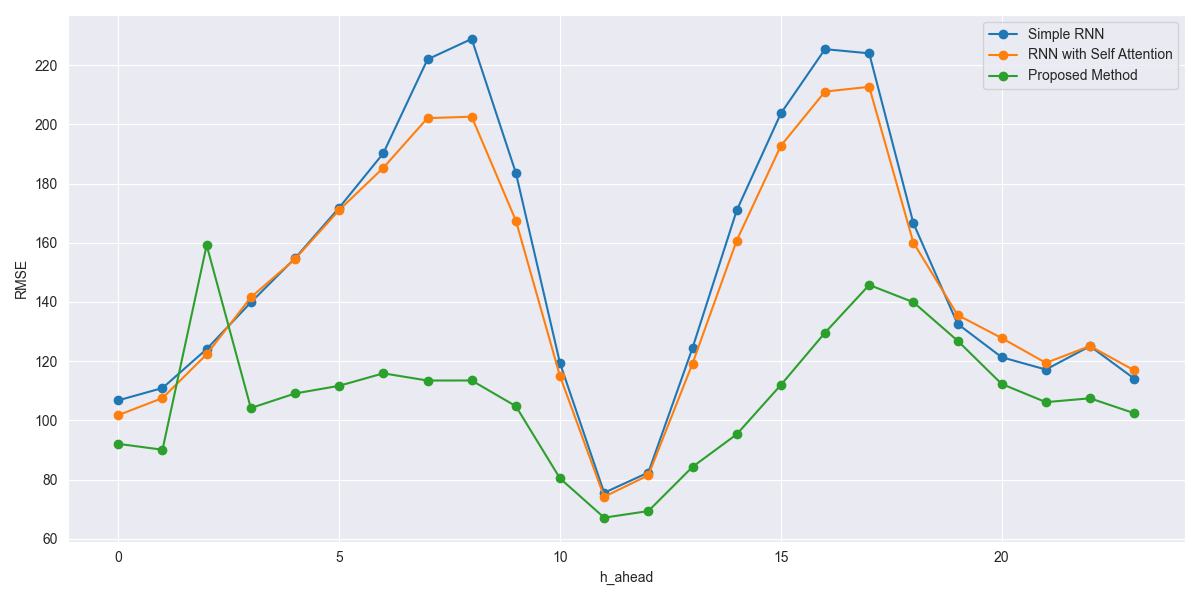}}
\caption{RNN ablations across four territories. Self-attention improves intra-day dependency modeling,
while adding Fourier seasonality further reduces RMSE and flattens the hourly error profiles.}
\label{fig:rmse_rnn_all}
\end{figure}

Figures~\ref{fig:rmse_rnn_all} show the results of comparison within the RNN structure. Self-attention consistently lowers RMSE relative to the plain RNN, especially around morning and evening peaks.
Adding Fourier seasonality further reduces error and flattens the profile, confirming that explicit harmonic features
enhance both interpretability and temporal generalization.

\begin{figure}[t]
\centering
\subfloat[AEP]{\includegraphics[width=0.48\linewidth]{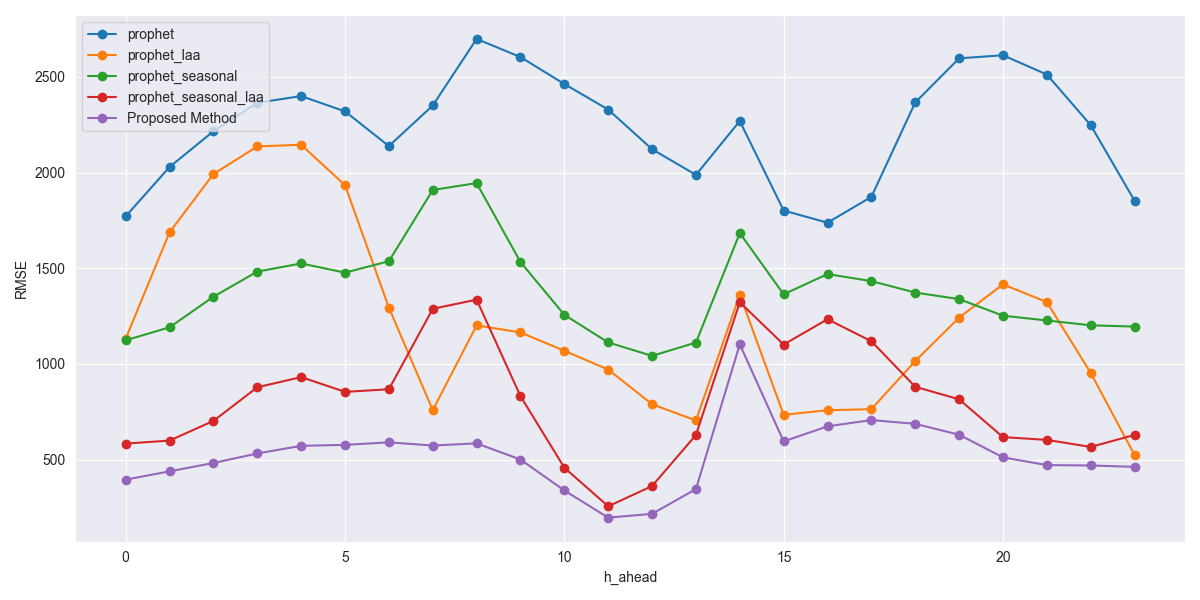}}
\subfloat[COMED]{\includegraphics[width=0.48\linewidth]{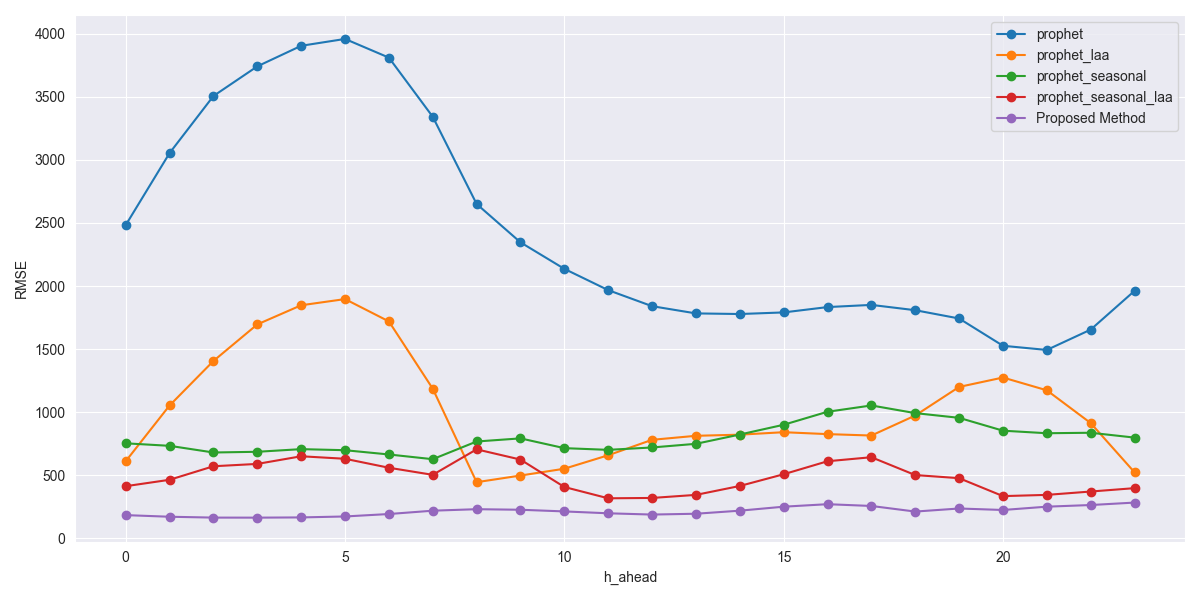}}\\[-2pt]
\subfloat[DAYTON]{\includegraphics[width=0.48\linewidth]{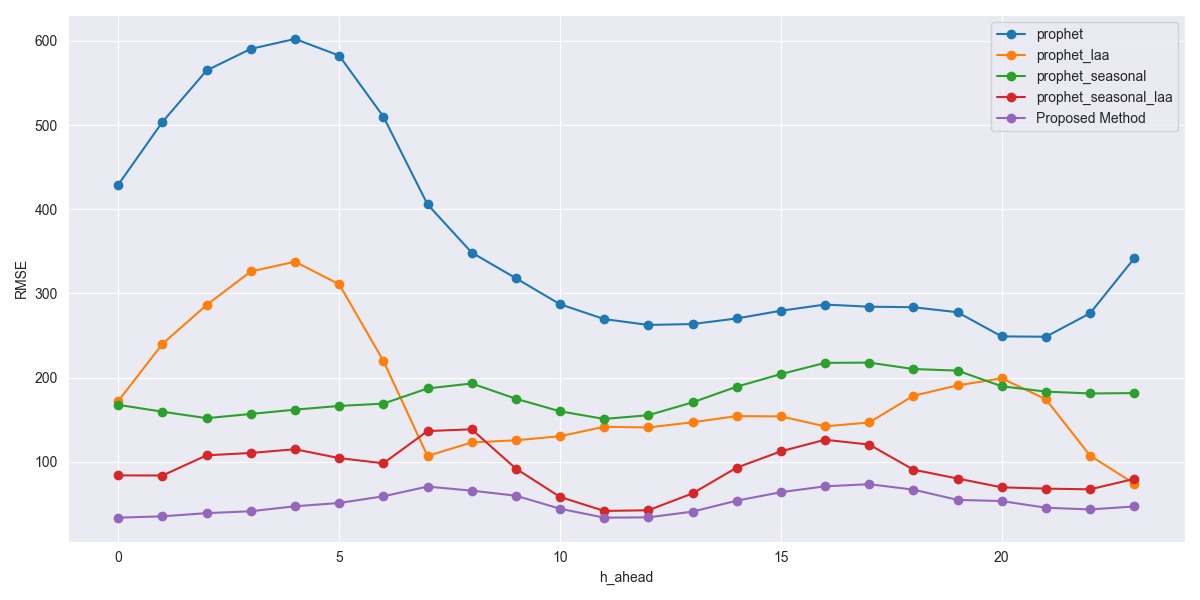}}
\subfloat[DEOK]{\includegraphics[width=0.48\linewidth]{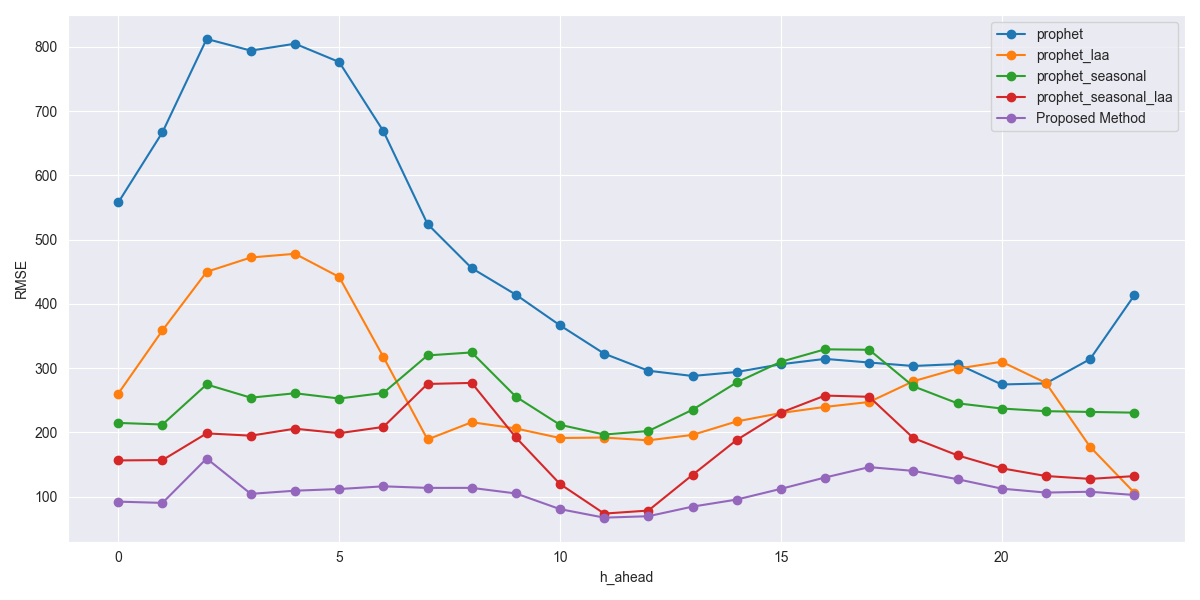}}
\caption{Prophet baselines versus Fourier RNN (ours; note the different $y$ scale from Figure \ref{fig:rmse_rnn_all}). Prophet variants include daily/weekly/yearly seasonalities
and optional LAA adjustment. Fourier RNN achieves the best mean RMSE and flattest hourly profiles.}
\label{fig:rmse_prophet_all}
\end{figure}

Figures~\ref{fig:rmse_prophet_all} compare across modeling families. While Prophet with LAA improves coherence across hours,
our proposed method achieves substantially lower RMSE and nearly flat horizon profiles,
demonstrating stronger adaptation to temporal cycles and trend variations.

Reading Figs.~\ref{fig:rmse_rnn_all} and~\ref{fig:rmse_prophet_all} jointly, the Simple RNN, even without explicit attention or harmonic inputs, performs competitively with Prophet+LAA, indicating a strong recurrent inductive bias for short-horizon dynamics. Incorporating Fourier seasonality further achieves the lowest RMSE and flattest horizon profiles, highlighting enhanced accuracy, uniformity, and more precise separation between trend and intra-day structure.

\subsection{Calibration of Predictive Uncertainty} \label{ss:calibration}

We evaluate uncertainty calibration by examining the empirical coverage of residual-based prediction intervals, which is expected to contain the actual value with probability $(1-\alpha)$.  
We compute the $r_h = \frac{n_{\text{rejected}}(h)}{n_{\text{total}}(h)}$ where $n_{\text{rejected}}(h)$ counts test windows whose actual value lies outside the interval.  
A well-calibrated model should have $r_h \approx \alpha$ across hours. Table ~\ref{tab:rejection} shows max, min, and mean of $r_h$

\begin{table}[ht]
\centering
\caption{Summary of rejection rates $r_h$ over 70 rolling test windows (95\% prediction interval).}
\begin{tabular}{lcccc}
\toprule
\textbf{Statistic} & \textbf{AEP} & \textbf{COMED} & \textbf{DAYTON} & \textbf{DEOK} \\
\midrule
Mean & 0.144 & 0.021 & 0.051 & 0.096 \\
Max  & 0.243 & 0.071 & 0.129 & 0.186 \\
Min  & 0.014 & 0.000 & 0.014 & 0.029 \\
\bottomrule
\end{tabular}
\label{tab:rejection}
\end{table}

For \textit{COMED} and \textit{DAYTON}, the mean rejection rates remain below or at the nominal level, indicating conservative coverage and stable uncertainty estimation.  
\textit{AEP} exhibits higher rejection, while \textit{DEOK} shows rejection near 0.096—somewhat above the nominal 0.05—implying moderate undercoverage.
Overall, these results suggest that residual-based Gaussian intervals provide a reasonable approach to quantifying uncertainty, with differences primarily reflecting dataset-specific volatility.

\subsection{Yearly-to-Hourly Downscaling Experiments}

To evaluate the proposed RNN-based structure on the yearly-to-hourly task, four models are compared: 
Prophet with yearly, monthly, and weekly seasonality; Prophet with the same seasonality and additional 
Linear Accuracy Adjustment (LAA); a two-stage hierarchical RNN (year~$\rightarrow$~day~$\rightarrow$~hour); 
and an RNN-augmented Prophet model that refines coarse-scale forecasts using the proposed RNN. 

\begin{figure}[t]
\centering
\subfloat[AEP]
{\includegraphics[width=0.48\linewidth]{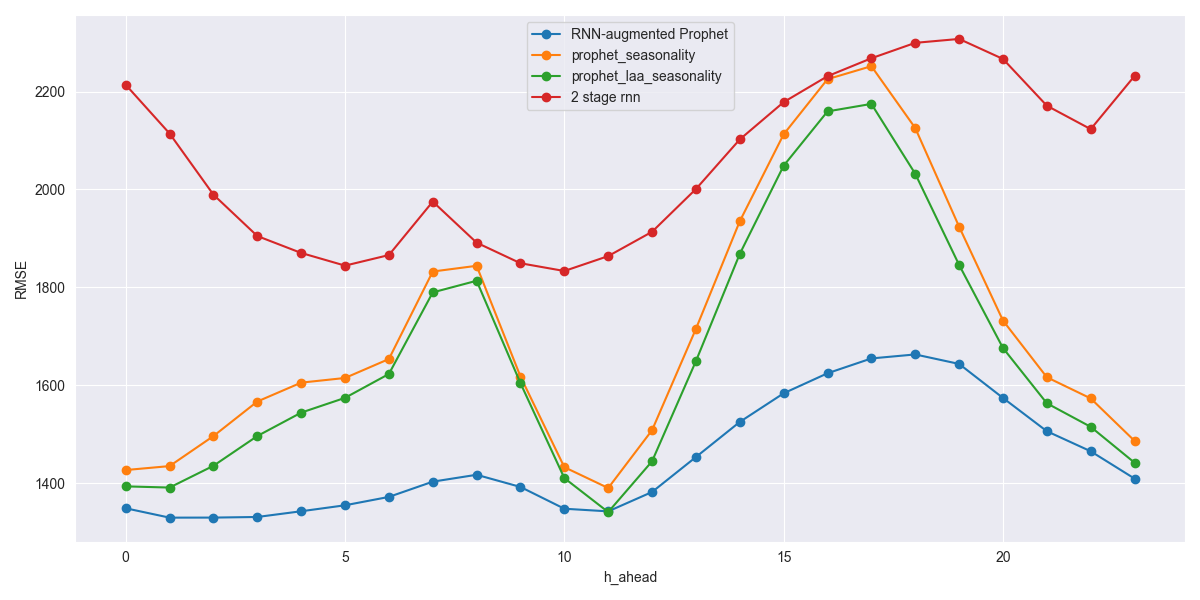}}
\subfloat[COMED]
{\includegraphics[width=0.48\linewidth]{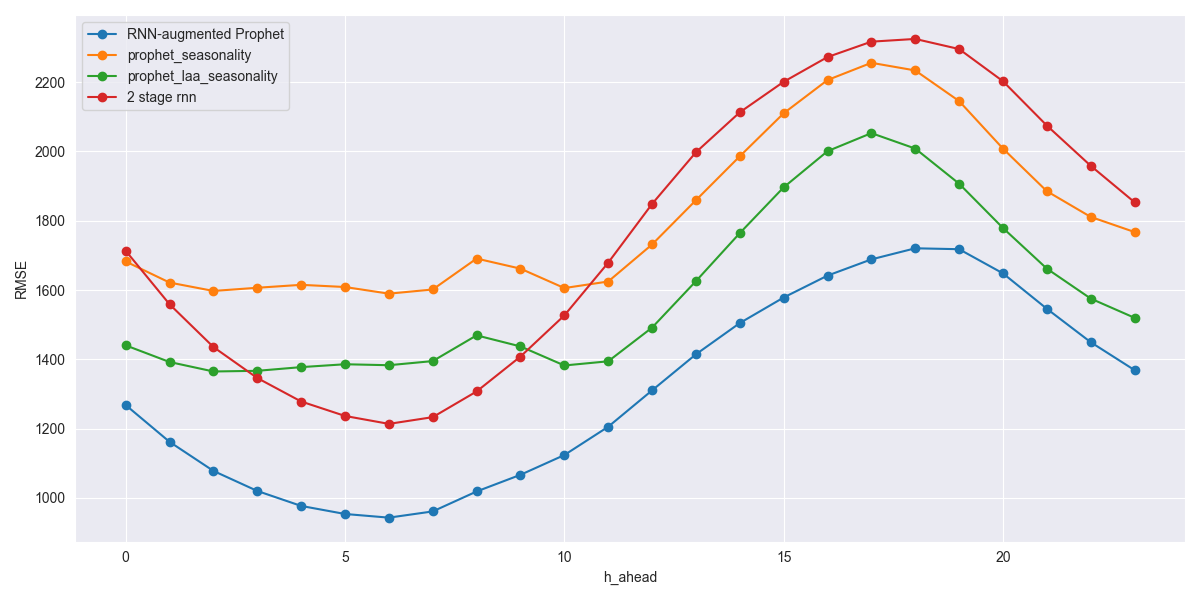}}\\[-2pt]
\subfloat[DAYTON]
{\includegraphics[width=0.48\linewidth]{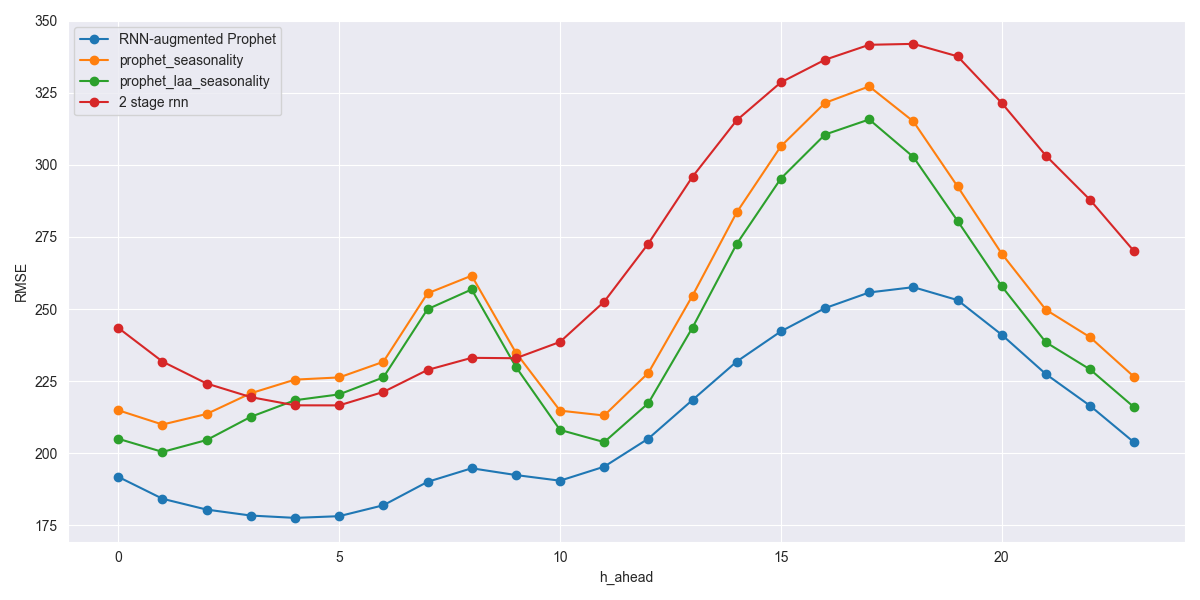}}
\subfloat[DEOK]
{\includegraphics[width=0.48\linewidth]{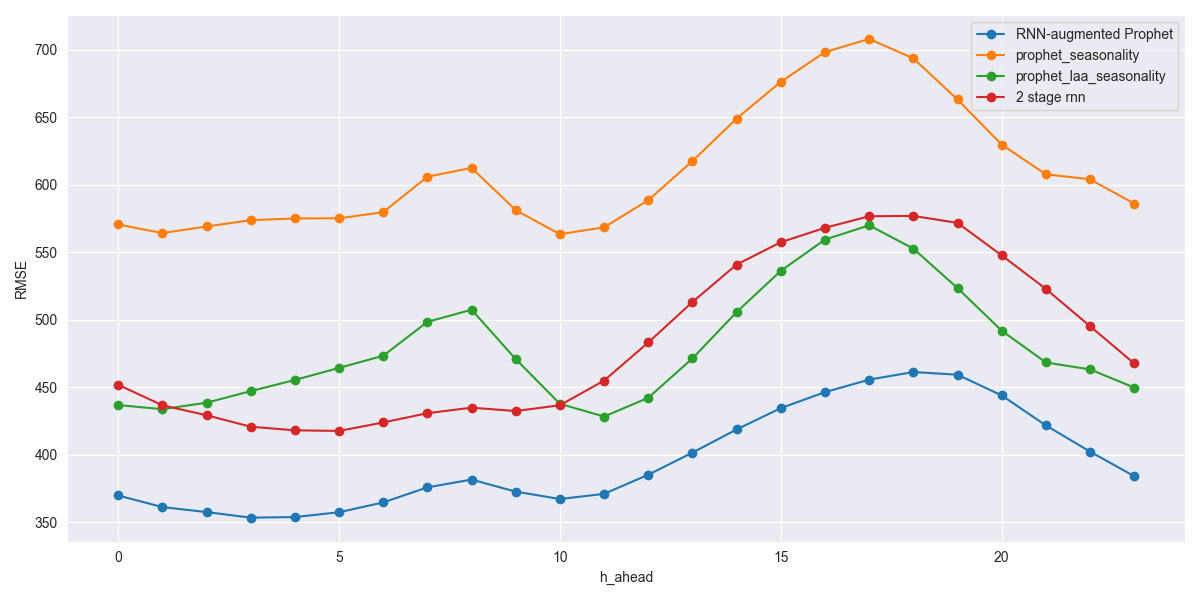}}
\caption{Yearly-to-hourly downscaling RMSE.}
\label{fig:yearly_rmse}
\end{figure}

Figure~\ref{fig:yearly_rmse} reports RMSE grouped by hourly positions (0--23). 
Prophet+seasonality and Prophet+seasonality+LAA reproduce broad seasonal patterns but tend to underfit short-term variability. 
The two-stage RNN captures detailed intra-year structure yet exhibits slightly higher RMSE due to limited 
training data, with less than ten years of samples available. In contrast, the RNN-augmented Prophet 
mitigates this limitation by coupling a coarse-scale forecaster with an RNN refiner---the Prophet component 
preserves interpretable long-term seasonality, while the RNN enhances intra-day variation. 

Overall, the RNN-augmented approach achieves the lowest RMSE and flattest hourly curves, demonstrating 
its ability to reconstruct fine-scale variability while maintaining consistent yearly-scale coherence.

\subsection{Reproducibility}
All experiments follow the same preprocessing, rolling splits, and evaluation pipeline.
Configuration files and random seeds are released for full reproducibility at 
\href{https://github.com/qic0-0/RNN_based_downscaling}{https://github.com/qic0-0/RNN\_based\_downscaling}.

\section{Conclusion}
The proposed RNN–Fourier architecture integrates a low-resolution recurrent path, Fourier-seasonal embeddings, and a timestep self-attention mechanism to capture complex temporal dependencies without explicit reconciliation. The model achieves accurate and stable horizon-wise downscaling performance across four PJM territories, outperforming Prophet-based and simplified RNN baselines. Beyond deterministic 
forecasting, the framework supports a probabilistic extension with residual-based uncertainty estimation and scales naturally to more challenging yearly-to-hourly downscaling tasks, demonstrating flexibility across both temporal and resolution hierarchies.

Future work includes extensions to spatial downscaling, which can be naturally carried out in our framework using a matrix data structure of size spatial sites by 24. While the Gaussian residual model yielded acceptable results in our predictive uncertainty calibration test using coverage values, an improvement is to use a generative formulation of the residual, albeit at the cost of a much more challenging  loss function. Finally, given the generality of our framework, we will experiment with other downscaling applications such as 
rainfall disaggregation, traffic intensity forecasting, or climate model refinement.

\section*{Acknowledgment}
    This material was based upon work
    supported by the U.S. Department of Energy, Office of Science,
    Office of Advanced Scientific Computing Research (ASCR) under
    Contract DE-AC02-06CH11347.

\bibliographystyle{IEEEtran}
\bibliography{reference}

@inproceedings{taylor2018prophet,
  title={Forecasting at Scale},
  author={Taylor, Sean J. and Letham, Benjamin},
  booktitle={Proceedings of the 23rd ACM SIGKDD International Conference on Knowledge Discovery and Data Mining},
  pages={813--821},
  year={2018}
}

@article{cho2014gru,
  title={Learning Phrase Representations using RNN Encoder–Decoder for Statistical Machine Translation},
  author={Cho, Kyunghyun and Merrienboer, Bart van and Gulcehre, Caglar and Bahdanau, Dzmitry and others},
  journal={EMNLP},
  year={2014}
}

@inproceedings{vaswani2017attention,
  title={Attention Is All You Need},
  author={Vaswani, Ashish and others},
  booktitle={NIPS},
  year={2017}
}

@article{ba2016layernorm,
  title={Layer Normalization},
  author={Ba, Jimmy Lei and Kiros, Jamie and Hinton, Geoffrey},
  journal={arXiv:1607.06450},
  year={2016}
}

@article{koutsoyiannis1996accurate,
  title     = {Simple Disaggregation by Accurate Adjusting Procedures},
  author    = {Koutsoyiannis, Demetris and Manetas, Alexandros},
  journal   = {Water Resources Research},
  volume    = {32},
  number    = {7},
  pages     = {2105--2117},
  year      = {1996},
  publisher = {American Geophysical Union},
  doi       = {10.1029/96WR00488}
}

@book{bracewell1999fourier,
  title={The Fourier Transform and Its Applications},
  author={Bracewell, Ronald N.},
  year={1999},
  publisher={McGraw-Hill}
}

@article{mystakidis2024energy,
  title={Energy forecasting: A comprehensive review of techniques and technologies},
  author={Mystakidis, Aristeidis and Koukaras, Paraskevas and Tsalikidis, Nikolaos and Ioannidis, Dimosthenis and Tjortjis, Christos},
  journal={Energies},
  volume={17},
  number={7},
  pages={1662},
  year={2024},
  publisher={MDPI}
}

@Article{hobbs2010regions,
  author    = {Hobbs, Benjamin F and Hu, Ming-Che and Chen, Yihsu and Ellis, J Hugh and Paul, Anthony and Burtraw, Dallas and Palmer, Karen L},
  journal   = {IEEE Transactions on Power Systems},
  title     = {From regions to stacks: spatial and temporal downscaling of power pollution scenarios},
  year      = {2010},
  number    = {2},
  pages     = {1179--1189},
  volume    = {25},
  publisher = {IEEE},
}

@Book{national1992national,
  author    = {National Research Council and Division on Engineering and Physical Sciences and Commission on Engineering and Technical Systems and Commission on Behavioral and Committee on the National Energy Modeling System, Energy Engineering Board and Commission on Engineering and Technical Systems, Committee on National Statistics},
  publisher = {National Academies Press},
  title     = {The national energy modeling system},
  year      = {1992},
}

@Article{fattahi2020systemic,
  author    = {Fattahi, Amirhossein and Sijm, Jos and Faaij, Andr{\'e}},
  journal   = {Renewable and Sustainable Energy Reviews},
  title     = {A systemic approach to analyze integrated energy system modeling tools: A review of national models},
  year      = {2020},
  pages     = {110195},
  volume    = {133},
  publisher = {Elsevier},
}

@Article{saravanan2013solution,
  author    = {Saravanan, Balasubramanian and Das, Siddharth and Sikri, Surbhi and Kothari, Dwarkadas Pralhaddas},
  journal   = {Frontiers in Energy},
  title     = {A solution to the unit commitment problem—a review},
  year      = {2013},
  number    = {2},
  pages     = {223--236},
  volume    = {7},
  publisher = {Springer},
}

@Article{xia2010optimal,
  author    = {Xia, X and Elaiw, AM},
  journal   = {Electric power systems research},
  title     = {Optimal dynamic economic dispatch of generation: A review},
  year      = {2010},
  number    = {8},
  pages     = {975--986},
  volume    = {80},
  publisher = {Elsevier},
}

@TechReport{hale2018demand,
  author      = {Hale, Elaine and Horsey, Henry and Johnson, Brandon and Muratori, Matteo and Wilson, Eric and Borlaug, Brennan and Christensen, Craig and Farthing, Amanda and Hettinger, Dylan and Parker, Andrew and others},
  institution = {National Renewable Energy Lab.(NREL), Golden, CO (United States)},
  title       = {The demand-side grid (dsgrid) model documentation},
  year        = {2018},
}

@Article{michel2021climate,
  author    = {Michel, Adrien and Sharma, Varun and Lehning, Michael and Huwald, Hendrik},
  journal   = {International Journal of Climatology},
  title     = {Climate change scenarios at hourly time-step over Switzerland from an enhanced temporal downscaling approach},
  year      = {2021},
  number    = {6},
  pages     = {3503--3522},
  volume    = {41},
  publisher = {Wiley Online Library},
}

\clearpage

\vspace{0.1cm}
\begin{flushright}
	\scriptsize \framebox{\parbox{2.5in}{(will be removed at publication; not part of page count) Government License: The
			submitted manuscript has been created by UChicago Argonne,
			LLC, Operator of Argonne National Laboratory (``Argonne").
			Argonne, a U.S. Department of Energy Office of Science
			laboratory, is operated under Contract
			No. DE-AC02-06CH11357.  The U.S. Government retains for
			itself, and others acting on its behalf, a paid-up
			nonexclusive, irrevocable worldwide license in said
			article to reproduce, prepare derivative works, distribute
			copies to the public, and perform publicly and display
			publicly, by or on behalf of the Government. The Department of Energy will provide public access to these results of federally sponsored research in accordance with the DOE Public Access Plan. http://energy.gov/downloads/doe-public-access-plan. }}
	\normalsize
\end{flushright}

\end{document}